%% file: main.tex
\documentclass[10pt,twocolumn,letterpaper]{article}

\usepackage{cvpr}              %

\input{preamble}
\definecolor{cvprblue}{rgb}{0.21,0.49,0.74}
\usepackage[pagebackref,breaklinks,colorlinks,allcolors=cvprblue]{hyperref}
\usepackage{multicol,multirow,booktabs}
\usepackage[table]{xcolor}
\usepackage{pifont}
\newcommand{\cmark}{\ding{51}}
\newcommand{\xmark}{\ding{55}}
\usepackage{caption}
\def\confName{CVPR}
\def\confYear{2027}

\title{3D CoCa v3: Contrastive Learners with Scene Graphs for\\Urban Scene Understanding}

\author{Nannan Zhang$^{1*}$~~
Ting Huang$^{2*}$~~
Zeyu Zhang$^{2*\dag}$~~
Haoyu Chen$^{3}$~~
Yanjie Zhu$^{1\ddag}$~~
Hao Tang$^{2\ddag}$\\[0.5em]
$^1$SIAT, CAS~~
$^2$School of Computer Science, Peking University~~
$^{3}$University of Oulu
\\[0.4em]
\small $^*$Equal contribution.
$^\dag$Project lead.
$^\ddag$Corresponding authors: yj.zhu@siat.ac.cn, bjdxtanghao@gmail.com.
}

\begin{document}
\maketitle
\begin{abstract}
Understanding large-scale 3D urban scenes requires representations that capture both multi-scale geometric structures and complex spatial relationships.
However, existing 3D vision-language methods primarily emphasize global semantic alignment or object-level representations, leaving structured relational modeling underexplored within unified contrastive and generative frameworks.
In this paper, we present \textbf{3D CoCa v3}, a scene-graph-enhanced contrastive and generative framework for urban 3D scene understanding.
Our framework features a Pyramid Vision Encoder for multi-scale geometric representation learning and a language-model-based decoder with question-guided caption retrieval and visual cross-attention.
We further introduce MSG-based relational descriptions to enrich contrastive 3D-language alignment with structured spatial relationships.
To support semantic and relational supervision, we construct City3D-9.6K, a multi-view urban vision-language dataset with aligned visual-geometric observations, scene captions, and scene graphs.
Extensive experiments on City-3DQA demonstrate that 3D CoCa v3 achieves 88.35\% and 64.79\% overall top-1 accuracy on the sentence-wise and city-wise splits, surpassing 3D CoCa v2 by 20.88 and 3.57 percentage points, respectively.
Furthermore, 3D CoCa v3 achieves the best performance on GeoEval3D, demonstrating its effectiveness in urban spatial reasoning and geometric understanding.
Code:~\url{https://github.com/AIGeeksGroup/3DCoCav3}.
\end{abstract}

\section{Introduction}
Understanding large-scale 3D urban scenes is essential for spatial intelligence, with applications in 3D scene captioning~\cite{huang2026dc}, spatial reasoning~\cite{huang20253d,huang2026consispace}, and embodied navigation~\cite{huang2026mobilevla,liu2025nav,huang2026mobilevla2}.
Compared with indoor environments, urban scenes exhibit complex geometric structures, diverse object categories, and intricate spatial relationships across multiple scales.
Understanding such environments requires not only recognizing individual objects but also capturing their spatial configurations and semantic relationships.
Recent advances in vision-language learning have enabled models to associate 3D observations with natural-language descriptions, supporting tasks such as question answering, visual grounding, and scene captioning.
These developments underscore the importance of learning 3D representations that capture both geometric structures and spatial relationships in complex urban environments.

\begin{figure}[t]
    \centering
    \small
    \includegraphics[width=\linewidth]{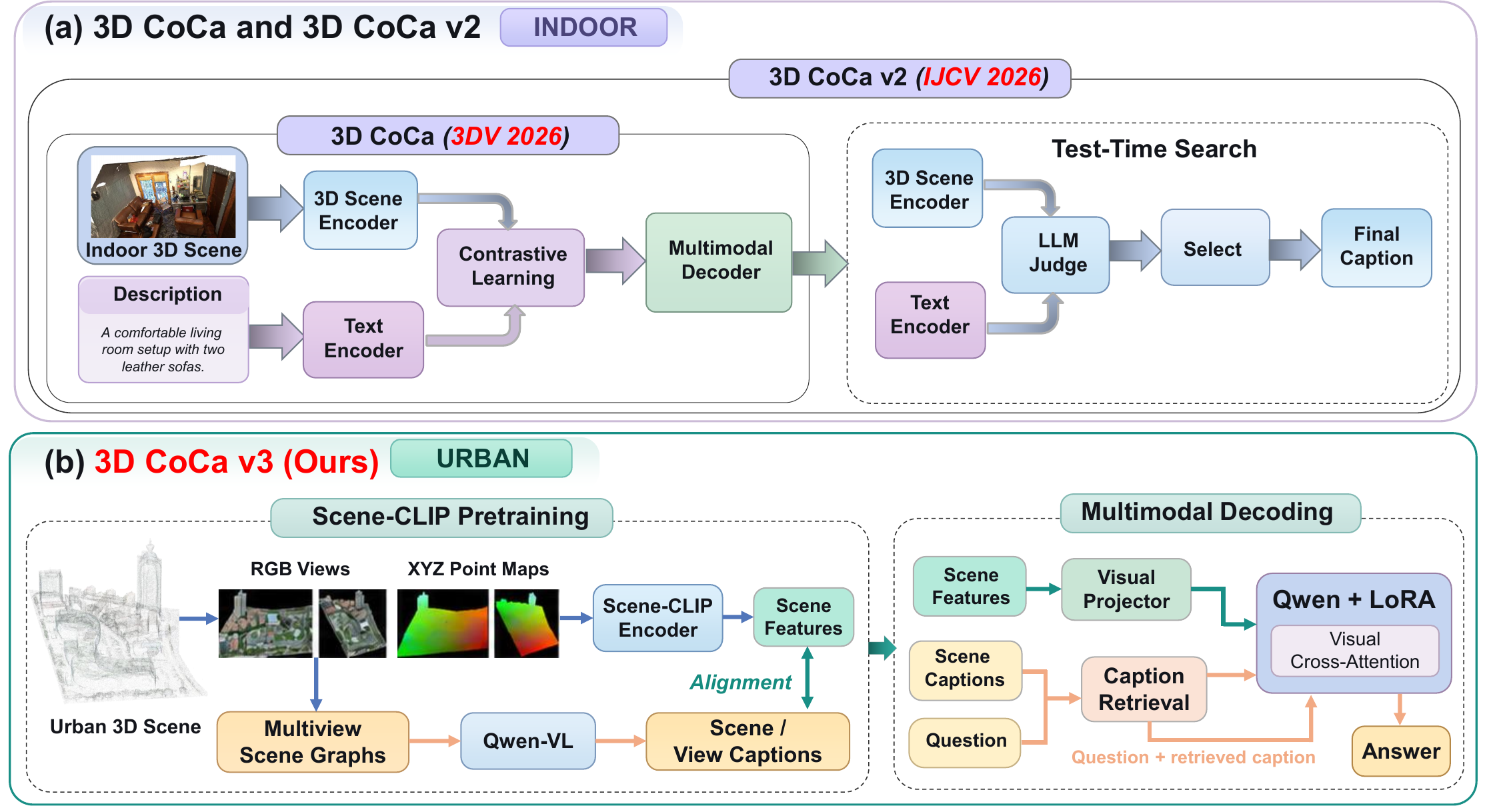}
    \caption{\textbf{Comparison of 3D CoCa, 3D CoCa v2, and \textcolor{red}{our 3D CoCa v3}.}
    Previous methods focus on indoor 3D scene understanding through contrastive learning, generative captioning, and test-time search.
    Our framework extends this paradigm to urban scenes with geometry-aware multi-view representations, MSG-based relational descriptions, and question-guided multimodal decoding.}
    \label{fig:teaser}
    \vspace{-0.6cm}
\end{figure}

\begin{table*}[t]
\centering
\caption{\textbf{Comparison of urban 3D vision-language datasets.}
City3D-9.6K provides aligned multi-view RGB images, point maps, and hierarchical language annotations.}
\label{tab:dataset_comparison}
\small
\setlength{\tabcolsep}{6pt}
\renewcommand{\arraystretch}{1.12}
\begin{tabular}{@{}llcccccc@{}}
\toprule
Dataset & Size & Multi-view & RGB--PM & View Cap. & Scene Cap. & Scene Graph & QA \\
\midrule
City-3DQA~\cite{sun2024city3dqa}
& 450K QA
& \xmark & \xmark & \xmark & \xmark
& \cmark & \cmark \\

SVM-City~\cite{sun2025city}
& 567K QA
& \cmark & \xmark & \xmark & \xmark
& \cmark & \cmark \\

3DCity-LLM-1.2M~\cite{chen20263dcity}
& 1.2M Samples
& \xmark & \xmark & \xmark & \cmark
& \cmark & \cmark \\
\midrule
\textbf{City3D-9.6K (Ours)}
& \textbf{9,650 Views}
& \cmark & \cmark & \cmark & \cmark
& \cmark & \cmark \\
\bottomrule
\end{tabular}
\vspace{-0.6cm}
\end{table*}

Existing 3D vision-language methods have evolved from task-specific understanding to unified representation learning and structured spatial reasoning.
Early approaches, such as ScanQA~\cite{azuma2022scanqa}, explored question answering over 3D scenes, while 3D-VLP~\cite{jin2023context} and 3D-VisTA~\cite{zhu2023vista} learned transferable representations through cross-modal alignment.
Building upon contrastive and generative learning, 3D CoCa~\cite{huang20263d} jointly optimizes 3D-text alignment and caption generation, while 3D CoCa v2~\cite{tang20263d} further enhances caption generalization through test-time search.
Beyond semantic alignment, scene-graph-based methods~\cite{liu20243d,zemskova20253dgraphllm} incorporate explicit object relationships to facilitate structured spatial reasoning.
Extending 3D understanding to urban environments, City-3DQA~\cite{sun2024city3dqa} introduces a city-scale question-answering benchmark and a scene-graph-enhanced reasoning framework, while 3DCity-LLM~\cite{chen20263dcity} explores coarse-to-fine representations for city-scale perception and understanding.
Collectively, these advances establish important foundations for geometric representation learning, cross-modal alignment, and relational reasoning in 3D environments.

Despite these advances, existing methods still face challenges in urban 3D scene understanding.
Current 3D vision-language frameworks primarily emphasize global semantic alignment or object-level representations, making it difficult to preserve fine-grained geometric details while capturing large-scale spatial context.
Meanwhile, although scene graphs provide explicit relational information, effectively integrating structured spatial relationships into joint contrastive and generative learning remains underexplored in complex urban environments.
These limitations motivate a unified 3D vision-language framework that jointly captures multi-scale geometric structures and spatial relationships through contrastive and generative representation learning.

To this end, we propose \textbf{3D CoCa v3}, a unified contrastive and generative framework for urban 3D scene understanding.
As illustrated in Fig.~\ref{fig:teaser}, our framework extends the 3D CoCa paradigm from indoor scenes to complex urban environments through geometry-aware representation learning and relational language supervision.
Our framework features a \emph{Pyramid Vision Encoder} for multi-scale geometric representation learning, coupled with a CapPa-inspired architecture and a larger language-model-based decoder for scene captioning and question answering.
We further develop \emph{MSG-based contrastive learning}, which incorporates structured scene-graph relationships into 3D-language alignment to enrich relational representations.
In addition, we construct \emph{City3D-9.6K}, a synthetic urban scene dataset with paired scene captions and scene graphs, providing complementary semantic and relational supervision.
Through joint contrastive and generative learning, 3D CoCa v3 integrates multi-scale geometry, structured spatial relationships, and natural-language semantics into a unified representation.
Extensive experiments on City-3DQA demonstrate the effectiveness of our framework, achieving \textbf{88.35\%} and \textbf{64.79\%} overall top-1 accuracy on the sentence-wise and city-wise splits, respectively, outperforming 3D CoCa v2 by \textbf{20.88} and \textbf{3.57} percentage points.
Furthermore, 3D CoCa v3 achieves the best performance on GeoEval3D, further demonstrating its capabilities in spatial reasoning and geometric understanding.

Our main contributions are summarized as follows:
\begin{itemize}
    \item We present \textbf{3D CoCa v3}, an enhanced contrastive and generative architecture featuring a Pyramid Vision Encoder, a CapPa-inspired captioning architecture, and a larger language-model-based decoder for multi-scale geometric representation learning and urban 3D scene understanding.
    \item We develop \textbf{MSG-based contrastive learning} that incorporates structured scene-graph relationships into 3D-language representation learning.
    We further introduce \textbf{City3D-9.6K}, a synthetic urban dataset with paired scene captions and scene graphs to support semantic and relational supervision.
    \item Extensive experiments on \textbf{City-3DQA} and \textbf{GeoEval3D} demonstrate the effectiveness of 3D CoCa v3.
    Our method achieves \textbf{88.35\%} and \textbf{64.79\%} overall top-1 accuracy on the sentence-wise and city-wise splits of City-3DQA, outperforming 3D CoCa v2 by \textbf{20.88} and \textbf{3.57} percentage points, respectively.
\end{itemize}

\section{Related Work}
\noindent\textbf{3D vision-language scene understanding.}
Early 3D vision-language methods, such as ScanQA~\cite{azuma2022scanqa}, 3D-VLP~\cite{jin2023context}, and 3D-VisTA~\cite{zhu2023vista}, established 3D question answering and cross-modal representation learning from point clouds.
SceneVerse~\cite{jia2024sceneverse} expanded grounded 3D-language supervision, while UniScene3D~\cite{mao2026rgb} learned geometry-aware representations from multi-view colored point maps.
Recent approaches further strengthen spatial representations through compact 3D proxies~\cite{proxy3d2026} and explicit 3D grounding~\cite{cheng2026gr3d}.
Beyond indoor environments, City-3DQA~\cite{sun2024city3dqa} establishes question answering over urban point clouds, whereas GeoProg3D~\cite{yasuki2025geoprog3d} supports compositional reasoning in city-scale 3D language fields.
These settings demand representations that capture both large-scale geometry and inter-object relations.

\begin{figure*}[!t]
    \centering
    \small
    \includegraphics[width=0.95\linewidth]{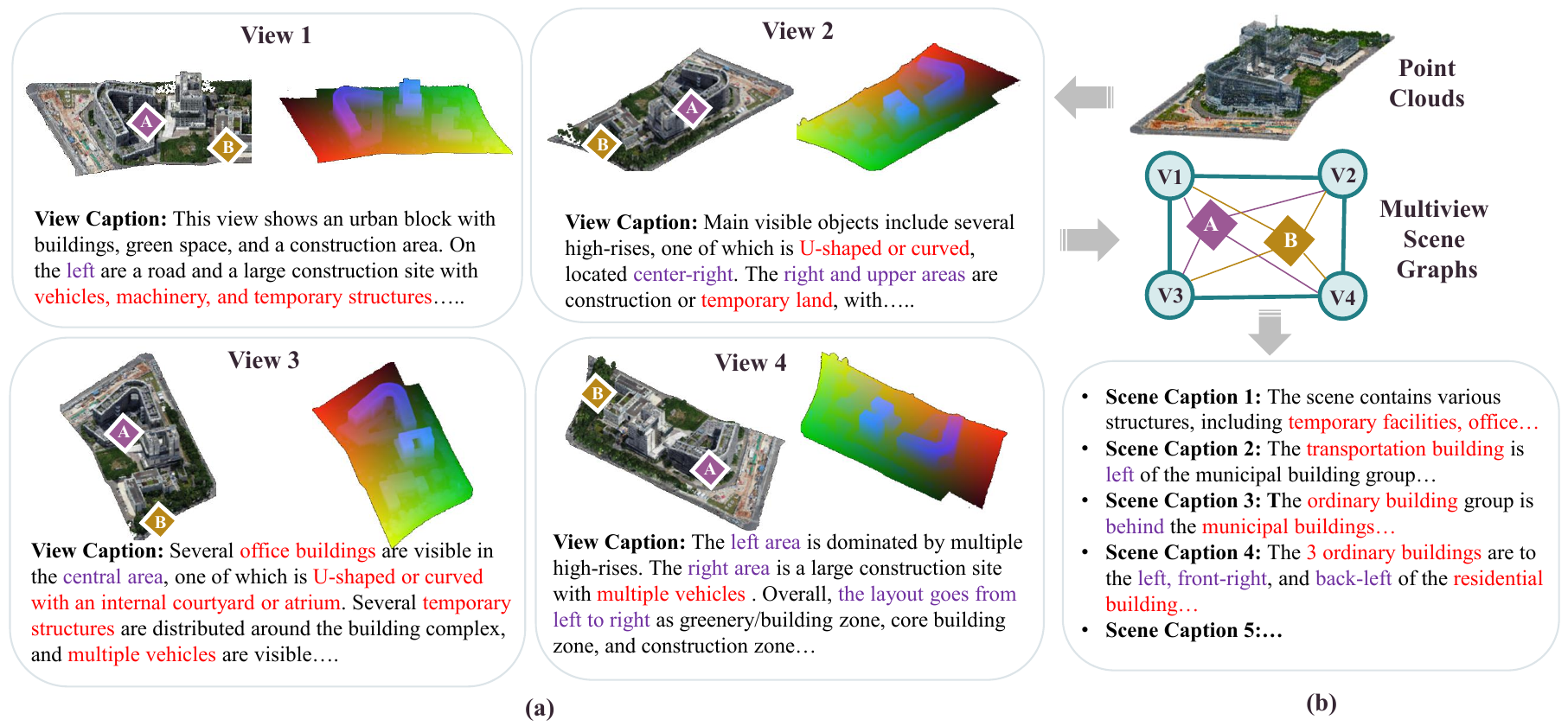}
    \caption{\textbf{Construction pipeline of City3D-9.6K.}
    (a) Multi-view RGB images and point maps are rendered from urban point clouds, with corresponding view-level captions generated by Qwen.
    (b) Multi-view scene graphs (MSG) associate shared entities and spatial relationships across viewpoints, providing structured context for scene-level caption generation.}
    \label{fig:city3d_pipeline}
    \vspace{-0.6cm}
\end{figure*}

\noindent\textbf{Scene graphs for spatial reasoning.}
Scene graphs organize objects and their semantic or spatial relationships into structured representations for compositional reasoning.
Sg-CityU~\cite{sun2024city3dqa} incorporates scene graphs into urban question answering, while 3DGraphLLM~\cite{zemskova20253dgraphllm} encodes relational structures as learnable inputs to language models.
Beyond object-centric relationships, MSG~\cite{zhang2024multiview} constructs topological scene graphs by associating place and object nodes across multiple viewpoints, while HSG~\cite{wang2026hsg} extends scene graph representations to hyperbolic space to capture hierarchical relationships.
View-on-Graph~\cite{liu2026view} further explores selective graph-based reasoning for zero-shot 3D visual grounding.
For cross-modal representation learning, CCL-3DSGG~\cite{chen2024clip} aligns 3D scene graph features with image and text representations, while scene graph-guided pre-training~\cite{liu20243d} leverages structured relations for fine-grained 3D-language contrastive alignment.
These advances highlight the potential of structured relational representations for geometry-aware vision-language understanding in complex 3D environments.

\noindent\textbf{Contrastive and generative 3D learning.}
CLIP~\cite{radford2021learning} established large-scale contrastive vision-language pre-training, while CoCa~\cite{yu2022coca} combined contrastive alignment with autoregressive captioning, and CapPa~\cite{tschannen2023image} explored parallel caption prediction.
For 3D captioning, Scan2Cap~\cite{chen2021scan2cap} and Vote2Cap-DETR++~\cite{chen2023end,chen2024vote2cap} model object descriptions and surrounding context; TOD$^3$Cap~\cite{todocap2025} extends dense captioning to outdoor scenes, and Curvature-Aware Captioning~\cite{he2026curvature} models geometric and semantic hierarchies through geodesic attention.
3D CoCa~\cite{huang20263d} transfers joint contrastive and generative learning to point clouds, with 3D CoCa v2~\cite{tang20263d} further introducing test-time search for caption selection.
Building on this line of work, we study graph-aware contrastive representations together with scene-level caption supervision and caption-conditioned question answering for urban 3D understanding.

\section{Dataset: City3D-9.6K}
\noindent\textbf{Dataset construction.}
We introduce \textbf{City3D-9.6K}, a multi-view 3D vision-language dataset built upon City-3DQA~\cite{sun2024city3dqa}.
The original benchmark comprises 193 urban point-cloud scenes with diverse geometric structures and spatial relationships.
As illustrated in Fig.~\ref{fig:city3d_pipeline}, we extend these scenes with multi-view visual observations and hierarchical language annotations.
Specifically, we render 50 viewpoints for each scene, yielding 9,650 RGB images and their corresponding point maps.
The paired RGB images and point maps provide aligned appearance and geometric information across different viewpoints, supporting geometry-aware visual representation learning.

\noindent\textbf{Hierarchical language annotations.}
We construct complementary language annotations at both view and scene levels.
For each rendered RGB image, we employ Qwen to generate a view-level caption describing visible objects, semantic attributes, and spatial configurations.
To capture spatial relationships beyond individual viewpoints, we further construct Multi-view Scene Graphs (MSG) that associate shared entities and their relationships across views, as shown in Fig.~\ref{fig:city3d_pipeline}(b).
The resulting graph structures provide relational context for Qwen-based scene caption generation, capturing the semantic organization and spatial relationships of the urban environment.
While view-level captions describe viewpoint-specific observations, MSG-based scene captions provide broader semantic context and structured relational information.
These complementary annotations support contrastive representation learning and generative scene understanding.

\begin{figure*}[htb]
    \centering
    \includegraphics[width=0.95\linewidth]{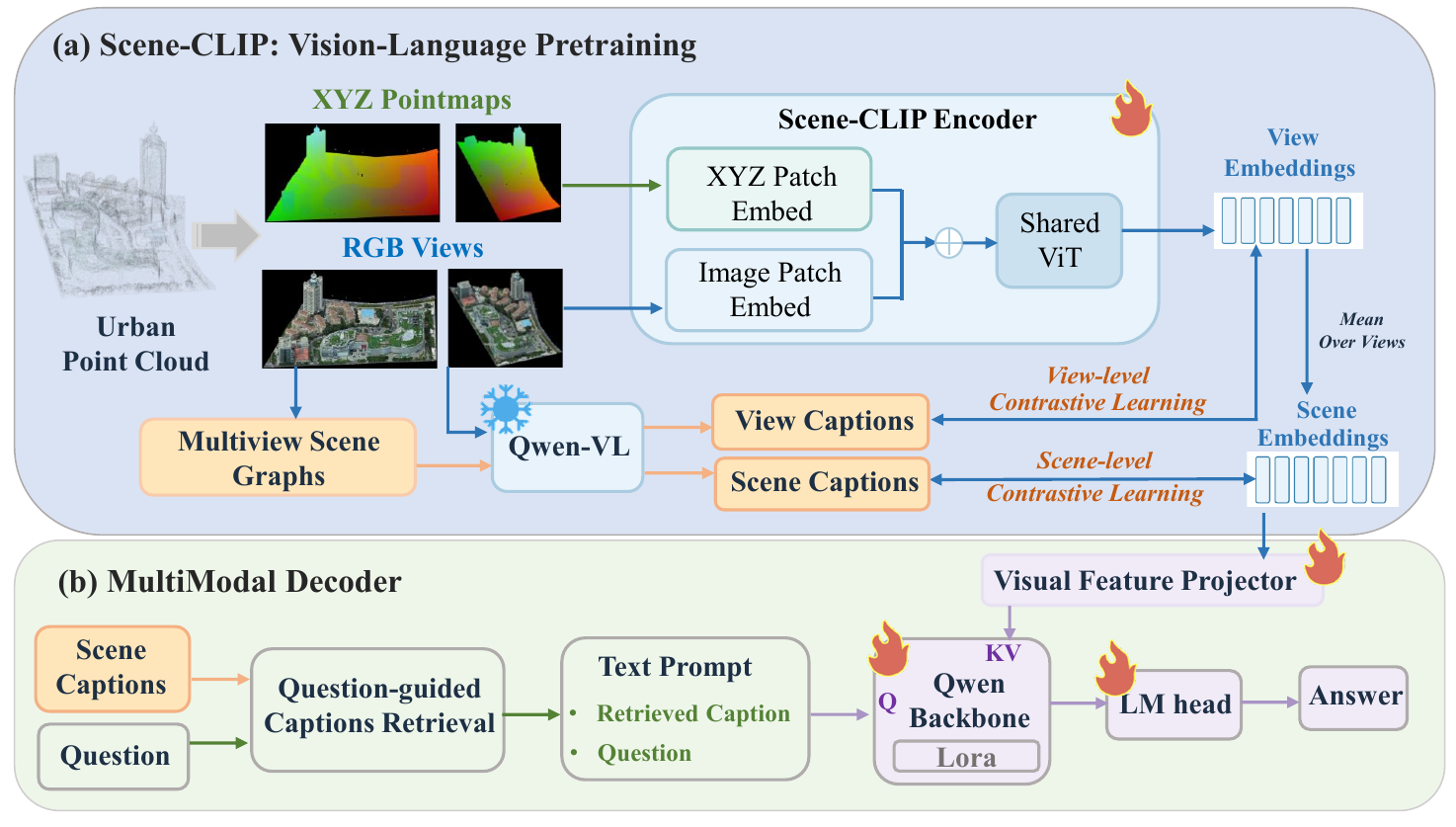}
    \caption{\textbf{Overview of 3D CoCa v3.}
    The framework learns geometry-aware representations from multi-view RGB images and XYZ point maps through Scene-CLIP contrastive pre-training, complemented by MSG-derived relational descriptions.
    A multimodal decoder subsequently integrates question-relevant scene captions and projected visual features into a LoRA-adapted Qwen backbone for urban 3D question answering.}
    \label{fig:overview}
    \vspace{-0.6cm}
\end{figure*}

\noindent\textbf{Dataset statistics and comparison.}
City3D-9.6K comprises 193 urban scenes, 9,650 rendered RGB images, 9,650 point maps, and 9,650 view-level captions, together with MSG annotations and scene-level captions.
The original question-answer pairs from City-3DQA are retained without introducing additional QA annotations.
Tab.~\ref{tab:dataset_comparison} compares City3D-9.6K with representative urban 3D vision-language datasets.
Our dataset extends the original urban point clouds with aligned multi-view visual, geometric, and linguistic observations, complemented by MSG-based scene-level descriptions.
This hierarchical annotation scheme provides semantic and relational supervision for urban 3D scene understanding.

\section{Method}

\subsection{Overview}
We present \textbf{3D CoCa v3}, a contrastive and generative framework for urban 3D scene understanding.
As illustrated in Fig.~\ref{fig:overview}, our framework consists of two stages: Scene-CLIP vision-language pre-training and multimodal language decoding.
The former learns geometry-aware representations through multi-view cross-modal alignment, while the latter integrates visual features and question-relevant scene descriptions for spatial question answering.

Given an urban point cloud $\mathcal{P}$, we render a set of multi-view observations $\mathcal{X}=\{(I_v,M_v)\}_{v=1}^{V}$, where $I_v$ and $M_v$ denote the RGB image and corresponding XYZ point map at viewpoint $v$.
During pre-training, a geometry-aware vision encoder learns view-level representations through contrastive alignment with visual and linguistic supervision, while scene-level representations are obtained by aggregating features across viewpoints.
Multi-view scene graphs (MSG) further provide structured relational information for generating scene descriptions.
During decoding, question-relevant captions are retrieved as textual context, and the pre-trained visual features are incorporated into a Qwen-based language model through cross-attention.
The resulting framework jointly exploits geometric observations and relational semantics for urban 3D scene understanding.

\subsection{Contrastive Representation Learning}
\noindent\textbf{Scene-CLIP vision encoder.}
We develop a geometry-aware Scene-CLIP encoder initialized from UniScene3D~\cite{mao2026rgb} to learn representations from aligned RGB images and XYZ point maps.
As shown in Fig.~\ref{fig:overview}(a), RGB and geometric observations are first processed by separate patch embedding layers.
The resulting embeddings are fused and passed through a shared vision transformer:
\begin{equation}
\begin{aligned}
    \mathbf{z}_v
    &=
    E_{\theta}
    \left(
    \mathcal{E}_{\mathrm{rgb}}(I_v)
    +
    \mathcal{E}_{\mathrm{xyz}}(M_v)
    \right),\\
    \mathbf{z}_s
    &=
    \frac{1}{V}\sum_{v=1}^{V}\mathbf{z}_v,
\end{aligned}
\label{eq:visual_encoding}
\end{equation}
where $\mathcal{E}_{\mathrm{rgb}}$ and
$\mathcal{E}_{\mathrm{xyz}}$ denote modality-specific patch embeddings, $E_{\theta}$ represents the shared visual encoder, and $\mathbf{z}_v\in\mathbb{R}^{d}$ is the view-level representation.
The scene embedding $\mathbf{z}_s$ aggregates information across viewpoints, providing a global representation of the urban environment.

\noindent\textbf{View- and scene-level contrastive alignment.}
We align the learned geometric representations with visual and linguistic information at complementary spatial granularities.
A frozen vision-language teacher provides RGB and text embeddings, denoted by $(\mathbf{r}_v,\mathbf{t}_v)$ at the view level and $(\mathbf{r}_s,\mathbf{t}_s)$ at the scene level.
We employ a symmetric InfoNCE objective, denoted by $\mathcal{L}_{\mathrm{NCE}}$, to maximize the similarity between matched cross-modal representations while distinguishing mismatched samples.

The multi-level alignment objective is formulated as
\begin{equation}
\begin{aligned}
\mathcal{L}_{\mathrm{align}}
={}&
\underbrace{
\mathcal{L}_{\mathrm{NCE}}(\mathbf{z}_v,\mathbf{r}_v)
+
\mathcal{L}_{\mathrm{NCE}}(\mathbf{z}_v,\mathbf{t}_v)
}_{\mathcal{L}_{\mathrm{view}}}\\
&+
\underbrace{
\mathcal{L}_{\mathrm{NCE}}(\mathbf{z}_s,\mathbf{r}_s)
+
\mathcal{L}_{\mathrm{NCE}}(\mathbf{z}_s,\mathbf{t}_s)
}_{\mathcal{L}_{\mathrm{scene}}}.
\end{aligned}
\label{eq:alignment_loss}
\end{equation}
The view-level objective preserves viewpoint-specific geometric and semantic information, while the scene-level objective encourages consistent representations of the overall urban environment.

Beyond batch-level alignment, we employ within-scene grounding to distinguish semantic descriptions associated with different viewpoints.
For a scene containing $V$ views, we construct a similarity matrix
\begin{equation}
    S_{vu}
    =
    s\,\hat{\mathbf{z}}_v^{\top}
    \hat{\mathbf{t}}_u,
    \label{eq:ground_similarity}
\end{equation}
where $\hat{\mathbf{z}}_v$ and $\hat{\mathbf{t}}_u$ are normalized view and text embeddings, respectively, and $s$ denotes the learned logit scale.
The corresponding bidirectional grounding loss is
\begin{equation}
    \mathcal{L}_{\mathrm{ground}}
    =
    \frac{1}{2}
    \left[
    \mathrm{CE}(\mathbf{S},\mathbf{y})
    +
    \mathrm{CE}(\mathbf{S}^{\top},\mathbf{y})
    \right],
    \label{eq:ground_loss}
\end{equation}
where $\mathbf{y}$ contains the diagonal matching indices.
This objective promotes fine-grained correspondence between geometric observations and viewpoint-specific language descriptions.

\noindent\textbf{Geometric neighborhood consistency.}
We additionally preserve geometric relationships among viewpoints through neighborhood consistency learning.
Given the Chamfer-distance-based neighborhood rankings, we construct a target distribution that combines nearest-neighbor assignments with soft neighborhood similarities:
\begin{equation}
    P_{vu}
    =
    \alpha\,\mathbb{1}[u=n(v)]
    +
    (1-\alpha)
    \frac{\exp(-d_{vu}/\tau_d)}
    {\sum_{k\neq v}\exp(-d_{vk}/\tau_d)},
    \label{eq:geo_target}
\end{equation}
where $n(v)$ denotes the nearest neighbor of view $v$, $d_{vu}$ is the geometric rank of view $u$ relative to $v$, and $\alpha$ controls the balance between hard and soft assignments.
The diagonal entries are excluded and the target distribution is normalized.

The geometric consistency objective is defined as
\begin{equation}
    \mathcal{L}_{\mathrm{geo}}
    =
    -\frac{1}{V}
    \sum_{v=1}^{V}
    \sum_{u\neq v}
    P_{vu}\log q_{vu},
    \label{eq:geo_loss}
\end{equation}
where $q_{vu}$ is the predicted similarity distribution obtained from normalized view embeddings.
The overall pre-training objective is
\begin{equation}
    \mathcal{L}_{\mathrm{pre}}
    =
    \mathcal{L}_{\mathrm{align}}
    +
    \mathcal{L}_{\mathrm{ground}}
    +
    \mathcal{L}_{\mathrm{geo}}.
    \label{eq:pretrain_loss}
\end{equation}
These complementary objectives encourage the learned representations to preserve cross-modal semantic correspondence and geometric consistency across viewpoints.

\subsection{Decoder}
\noindent\textbf{Question-guided caption retrieval.}
As illustrated in Fig.~\ref{fig:overview}(b), the decoder exploits scene descriptions generated from MSG-based relational information.
Given a question $Q$ and a set of candidate scene captions
$\mathcal{C}_s=\{C_1,\ldots,C_K\}$,
we employ a question-guided retrieval strategy to identify the most relevant description:
\begin{equation}
    C^{*}
    =
    \mathcal{R}(Q,\mathcal{C}_s),
    \label{eq:caption_retrieval}
\end{equation}
where $\mathcal{R}$ denotes a rule-based relevance function that considers entity correspondence, question type, and spatial relationships.
The retrieval mechanism selects a caption that provides supporting semantic or relational context.
When no candidate satisfies the relevance criteria, the decoder uses the question without additional caption context.

\noindent\textbf{Visual-language decoding.}
We employ a pretrained Qwen-based language model with parameter-efficient LoRA adaptation.
The Scene-CLIP encoder is frozen during this stage, and its multi-view representations serve as visual memory.
The retrieved caption $C^{*}$ is incorporated into the textual prompt together with the question:
\begin{equation}
    \mathbf{H}
    =
    D_{\phi}
    \left(
    [C^{*};Q]
    \right),
    \label{eq:text_encoding}
\end{equation}
where $D_{\phi}$ denotes the LoRA-adapted language backbone and $\mathbf{H}$ represents its textual hidden states.

To integrate geometric information, the pre-trained view representations are projected into the language-model hidden space:
\begin{equation}
    \mathbf{Z}
    =
    f_{\mathrm{proj}}
    \left(
    [\mathbf{z}_1,\ldots,\mathbf{z}_V]
    \right),
    \label{eq:visual_projection}
\end{equation}
where $f_{\mathrm{proj}}$ is a trainable visual feature projector.
The textual representations attend to the projected visual features through a gated cross-attention module:
\begin{equation}
\begin{aligned}
    \mathbf{A}
    &=
    \mathrm{CrossAttn}
    (\mathbf{H},\mathbf{Z},\mathbf{Z}),\\
    \mathbf{H}'
    &=
    \mathbf{H}
    +
    \gamma\,\mathrm{LN}(\mathbf{A}),
\end{aligned}
\label{eq:visual_cross_attention}
\end{equation}
where $\gamma$ is a learnable fusion parameter.
The fused hidden states are subsequently processed by the language-model head for answer prediction.

\begin{table*}[t]
\centering
\caption{\textbf{Comparison on City-3DQA~\cite{sun2024city3dqa}.}
Top-1 (Acc@1) and top-10 (Acc@10) accuracy (\%) are reported for single-hop, multi-hop, and all questions under sentence-wise and city-wise splits.}
\label{tab:city3dqa}
\setlength{\tabcolsep}{4pt}
\renewcommand{\arraystretch}{1.12}
\resizebox{\linewidth}{!}{%
\begin{tabular}{lc*{12}{c}}
\toprule
\multirow{3}{*}{\textbf{Method}}
& \multirow{3}{*}{\textbf{Input}}
& \multicolumn{6}{c}{\textbf{Sentence-wise}}
& \multicolumn{6}{c}{\textbf{City-wise}} \\
\cmidrule(lr){3-8}\cmidrule(lr){9-14}
& & \multicolumn{2}{c}{Single-hop}
& \multicolumn{2}{c}{Multi-hop}
& \multicolumn{2}{c}{All}
& \multicolumn{2}{c}{Single-hop}
& \multicolumn{2}{c}{Multi-hop}
& \multicolumn{2}{c}{All} \\
\cmidrule(lr){3-4}\cmidrule(lr){5-6}\cmidrule(lr){7-8}
\cmidrule(lr){9-10}\cmidrule(lr){11-12}\cmidrule(lr){13-14}
& & acc@1$\uparrow$ & acc@10$\uparrow$ & acc@1$\uparrow$ & acc@10$\uparrow$ & acc@1$\uparrow$ & acc@10$\uparrow$
& acc@1$\uparrow$ & acc@10$\uparrow$ & acc@1$\uparrow$ & acc@10$\uparrow$ & acc@1$\uparrow$ & acc@10$\uparrow$ \\
\midrule
Qwen-VL~\cite{bai2023qwenvl}
& Image
& 30.53 & 70.85 & 9.76 & 58.45 & 18.81 & 63.86
& 30.79 & 71.07 & 9.78 & 57.07 & 19.75 & 63.71 \\

LLaVA~\cite{liu2023visual}
& Image
& 33.93 & 77.02 & 10.33 & 59.92 & 20.60 & 67.37
& 32.56 & 76.94 & 9.84 & 58.07 & 20.56 & 67.02 \\

Qwen~\cite{bai2023qwen}
& Scene Graph
& 55.25 & 85.41 & 11.21 & 63.48 & 30.35 & 73.84
& 55.40 & 85.49 & 12.59 & 66.35 & 31.31 & 75.26 \\

Llama-2~\cite{touvron2023llama}
& Scene Graph
& 60.51 & 86.34 & 20.00 & 75.13 & 37.66 & 80.02
& 60.03 & 86.18 & 18.82 & 73.17 & 38.37 & 79.34 \\

ScanQA~\cite{azuma2022scanqa}
& Point Cloud
& 76.42 & 90.75 & 28.31 & 86.46 & 49.28 & 88.34
& 64.84 & 88.73 & 27.03 & 84.37 & 47.33 & 86.45 \\

CLIP-Guided~\cite{parelli2023clip}
& Point Cloud
& 74.54 & 98.49 & 33.73 & 97.54 & 51.55 & 98.38
& 63.05 & 98.35 & 32.41 & 97.12 & 46.94 & 98.00 \\

3D-VLP~\cite{jin2023context}
& Point Cloud
& 72.78 & 98.55 & 35.54 & 97.76 & 51.72 & 98.40
& 64.03 & 98.42 & 34.95 & 97.19 & 48.74 & 98.33 \\

3D-VisTA~\cite{zhu2023vista}
& Point Cloud
& 79.23 & 98.52 & 44.67 & 97.85 & 59.63 & 98.37
& 71.28 & 98.47 & 43.87 & 97.56 & 56.74 & 98.48 \\

Sg-CityU~\cite{sun2024city3dqa}
& Point Cloud + Scene Graph
& 80.95 & 98.86 & 50.75 & 98.66 & 63.94 & 98.81
& \textbf{78.46} & 98.76 & 50.50 & 98.45 & 63.76 & 98.68 \\

\midrule

\rowcolor{gray!10}
3D CoCa~\cite{huang20263d}  & Point Cloud  &80.10 &98.73 &50.43 &99.39 &63.49 &99.11  &72.99 &98.52 &48.67 & \textbf{99.55} &59.29 &99.10 \\

\rowcolor{gray!10}
3D CoCa v2~\cite{tang20263d}  & Point Cloud  &82.20 &98.76 &55.89 &\textbf{99.47} &67.47 &\textbf{99.16}  &75.73 &\textbf{99.05} &49.96 & 99.35&61.22 & \textbf{99.22}\\ 

\rowcolor{yellow!20}
\textbf{3D CoCa v3 (Ours)} & RGB + Pointmap + MSG & \textbf{97.73} & \textbf{99.42} & \textbf{80.97} & 96.52 &\textbf{88.35} &97.80 & 77.38 & 93.44 &\textbf{55.02}& 95.92 &\textbf{64.79} &94.84\\

\bottomrule
\end{tabular}%
}
\vspace{-0.5cm}
\end{table*}

\noindent\textbf{Answer generation.}
The decoder predicts the answer autoregressively, conditioned on the question, retrieved scene caption, and visual representations.
Given a target answer $A=(a_1,\ldots,a_T)$, the training objective is
\begin{equation}
    \mathcal{L}_{\mathrm{QA}}
    =
    -\sum_{t=1}^{T}
    \log p_{\Theta}
    \left(
    a_t\mid a_{<t},Q,C^{*},\mathbf{Z}
    \right),
    \label{eq:qa_loss}
\end{equation}
where $\Theta$ denotes the trainable decoder parameters.
The language-model backbone is adapted through LoRA, while the visual projector and cross-attention module are trained to establish geometric-language interactions.
This design enables question answering to exploit complementary information from multi-view geometric observations and structured scene descriptions.

\section{Experiments}
\label{sec:Experiment}

\subsection{Benchmark and Metrics}
\noindent\textbf{City-3DQA.}
We evaluate 3D CoCa v3 on City-3DQA~\cite{sun2024city3dqa}, a city-scale 3D question-answering benchmark comprising approximately 450K question-answer pairs across 193 scenes from six cities.
Following the official evaluation protocol, we consider two splits: \emph{sentence-wise}, where question-answer pairs are divided into training, validation, and test sets across all six cities, and \emph{city-wise}, where the splits contain disjoint cities.
The benchmark includes single-hop questions involving direct scene understanding and multi-hop questions requiring compositional spatial reasoning.
We report top-1 accuracy (Acc@1) and top-10 accuracy (Acc@10) for single-hop, multi-hop, and all questions.

\noindent\textbf{GeoEval3D.}
We further evaluate geographic spatial reasoning on GeoEval3D~\cite{yasuki2025geoprog3d}, a city-scale benchmark introduced by GeoProg3D.
GeoEval3D comprises 952 queries across five tasks: grounding (GRD), spatial reasoning (SPR), comparison (CMP), counting (CNT), and measurement (MES).
Following the evaluation protocol, we report accuracy for SPR and CMP, mean absolute error (MAE) for CNT, and measurement errors in meters for height (MES-H) and distance (MES-D).
We evaluate on both GoogleEarth and UrbanScene3D, with CMP and MES-H excluded from UrbanScene3D following the original benchmark.
Higher accuracy and lower MAE indicate better performance.

\subsection{Implementation Details}
We initialize the visual encoder from UniScene3D~\cite{mao2026rgb} and perform contrastive pre-training on the RGB--point-map observations from City3D-9.6K.
Each RGB image and its corresponding point map are processed at a resolution of $224\times224$, with 32 views sampled per scene during pre-training.
The encoder is optimized using AdamW with an initial learning rate of $1\times10^{-4}$ and a warmup cosine learning-rate schedule.
For downstream question answering, we extract and cache visual representations from 50 rendered views per scene.

We investigate CapPa- and CoCa-style decoder architectures with pretrained Qwen-based language models.
Our final decoder uses a 2B-parameter Qwen backbone with LoRA adaptation and visual cross-attention.
The language-model backbone is frozen, while the LoRA parameters, visual projection layers, and cross-attention module are optimized.
We use a LoRA rank of 16, an alpha of 32, and eight attention heads for visual cross-attention.
The decoder is trained with AdamW using a learning rate of $1\times10^{-6}$ for 30 epochs.
For caption-conditioned question answering, a question-relevant scene description is selected from the candidate captions and incorporated into the language prompt, while the visual representations are provided through cross-attention.

\begin{table*}[t]
\centering
\caption{
\textbf{Comparison on GeoEval3D~\cite{yasuki2025geoprog3d}.}
We report accuracy (\%) for spatial reasoning (SPR) and comparison (CMP), and mean absolute error (MAE) for counting (CNT) and measurement (MES-H/MES-D). Measurement errors are in meters.}
\label{tab:geoeval3d}
\setlength{\tabcolsep}{5pt}
\renewcommand{\arraystretch}{1.1}
\resizebox{0.8\textwidth}{!}{
\begin{tabular}{lccccc|ccc}
\toprule
& \multicolumn{5}{c|}{GoogleEarth} & \multicolumn{3}{c}{UrbanScene3D} \\
\cmidrule(lr){2-6} \cmidrule(lr){7-9}
Method & SPR$\uparrow$ & CMP$\uparrow$ & CNT$\downarrow$ & MES-H$\downarrow$ & MES-D$\downarrow$ & SPR$\uparrow$ & CNT$\downarrow$ & MES-D$\downarrow$ \\
& Acc. (\%) & Acc. (\%) & MAE & MAE (m) & MAE (m) & Acc. (\%) & MAE & MAE (m) \\ \midrule
Qwen2.5-VL-7B~\cite{qwen25vl}           & 53.39 & 26.95 & 2.65 & 59.68 & 175.44 & 47.24 & 4.00 & 412.86 \\
InternVL2.5-8B~\cite{chen2024internvl}  & 54.27 & 26.95 & 2.79 & 51.30 & 157.14 & 52.47 & 4.23 & 318.71 \\
GeoChat~\cite{kuckreja2024geochat}      & 57.23 & 41.99 & 2.89 & 84.74 & 89.34  & 56.76 & 3.69 & 328.68 \\
LHRS-BOT~\cite{muhtar2024lhrs}          & 49.45 & 27.52 & 4.85 & 46.17 & 104.49 & 41.94 & 3.49 & 438.94 \\
VHM~\cite{pang2025vhm}                   & 54.55 & 39.82 & 5.28 & 52.58 & 135.50 & 56.92 & 4.34 & 354.91 \\
TEOChat~\cite{irvin2025teochat}          & 59.04 & 48.11 & 2.84 & 150.39 & 198.89& 57.99 & 4.06 & 359.71 \\
GeoProg3D~\cite{yasuki2025geoprog3d}     & 64.00 & 59.73 & 2.00 & 45.24 & \textbf{49.28}  & 60.87 & 2.51 & 139.51 \\
\midrule
\rowcolor{gray!10}
3D CoCa~\cite{huang20263d} & 44.44 & 75.00 & 2.67 & 91.67 & 129.33 & 50.14 & 1.38 & 191.87 \\
\rowcolor{gray!10}
3D CoCa v2~\cite{tang20263d} & 59.44 & 50.00 & 2.03 & 73.34 & 98.63 & 59.23 & 1.13 & 180.45 \\
\rowcolor{yellow!20}
\textbf{3D CoCa v3 (Ours)}
& \textbf{75.56} & \textbf{82.34} & \textbf{1.67} & \textbf{40.38} & 55.34
& \textbf{68.33} & \textbf{0.50} & \textbf{135.67} \\
\bottomrule
\end{tabular}
}
\vspace{-0.6cm}
\end{table*}

\subsection{Main Results}
\noindent\textbf{Comparison on City-3DQA.}
As shown in Tab.~\ref{tab:city3dqa}, 3D CoCa v3 achieves the highest overall Acc@1 under both evaluation splits.
Under the sentence-wise split, our method achieves \textbf{88.35\%} overall Acc@1, outperforming 3D CoCa v2 by \textbf{20.88} percentage points.
Notably, single-hop and multi-hop accuracies improve from 82.20\% and 55.89\% to \textbf{97.73\%} and \textbf{80.97\%}, respectively, with a substantial gain of \textbf{25.08} percentage points on multi-hop questions.
Under the more challenging city-wise split, our method achieves \textbf{64.79\%} overall Acc@1, surpassing 3D CoCa v2 and Sg-CityU by \textbf{3.57} and \textbf{1.03} percentage points, respectively.
It also achieves the highest multi-hop accuracy of \textbf{55.02\%}, outperforming 3D CoCa v2 by \textbf{5.06} percentage points.
Although Acc@10 remains lower than that of several baselines, the consistent gains in overall and multi-hop Acc@1 highlight the effectiveness of our framework in top-ranked answer prediction and compositional spatial reasoning.

\noindent\textbf{Comparison on GeoEval3D.}
As shown in Tab.~\ref{tab:geoeval3d}, 3D CoCa v3 achieves the best performance on seven of eight evaluated metrics.
On GoogleEarth, our method achieves \textbf{75.56\%} spatial reasoning accuracy and \textbf{82.34\%} comparison accuracy, surpassing GeoProg3D by \textbf{11.56} and \textbf{22.61} percentage points, respectively.
It also achieves lower counting and height estimation errors of \textbf{1.67} and \textbf{40.38}\,m, while remaining competitive in distance estimation.
On UrbanScene3D, our method outperforms all compared approaches across all three metrics, achieving \textbf{68.33\%} spatial reasoning accuracy, \textbf{0.50} counting MAE, and \textbf{135.67}\,m distance MAE.
These results demonstrate the effectiveness of 3D CoCa v3 in both relational and metric spatial reasoning.

\begin{figure*}[htb]
    \centering
    \includegraphics[width=\linewidth]{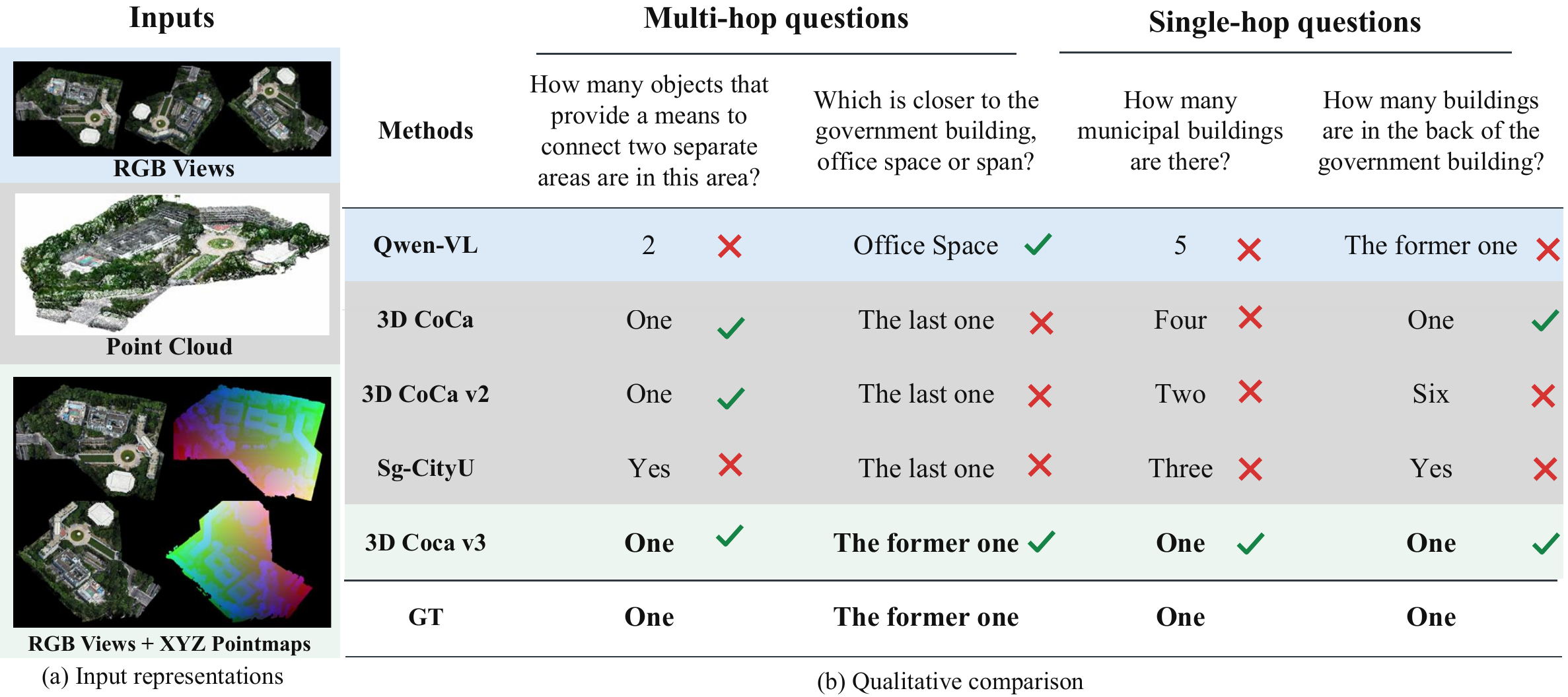}
    \caption{\textbf{Qualitative comparison on City-3DQA under the city-wise split.}
    We compare 3D CoCa v3 with representative baselines on single-hop and multi-hop questions involving object counting and spatial relationships.
    Our method correctly answers all four questions in the illustrated unseen urban scene.}
    \label{fig:qualitative}
    \vspace{-0.6cm}
\end{figure*}

\subsection{Ablation Study}

\noindent\textbf{Effect of contrastive learning objectives.}
Tab.~\ref{tab:loss_ablation} investigates the contributions of view-level and scene-level alignment during contrastive pre-training.
Removing view-level alignment decreases overall Acc@1 from 64.79\% to 62.05\%, while removing scene-level alignment results in a drop of 1.37 percentage points.
Both variants exhibit degraded performance on multi-hop questions, with accuracy decreasing by 2.78 and 1.74 percentage points, respectively.
These results highlight the complementary roles of fine-grained view-level alignment and holistic scene-level alignment in learning geometry-aware representations for urban spatial reasoning.

\begin{table}[bhtp]
\vspace{-0.3cm}
\small
\centering
\caption{\textbf{Effect of contrastive learning objectives.}
Ablation of view-level and scene-level alignment during contrastive pre-training on City-3DQA under the city-wise split.
Acc@1 (\%) is reported.}
\label{tab:loss_ablation}
\small
\setlength{\tabcolsep}{4pt}
\renewcommand{\arraystretch}{1.12}
\begin{tabular}{@{}lccc@{}}
\toprule
Configuration & Single-hop & Multi-hop & All \\
\midrule
w/o View-level Alignment
& 74.70 & 52.24 & 62.05 \\
w/o Scene-level Alignment
& 76.47 & 53.28 & 63.42 \\
\midrule
Full Model
& \textbf{77.38} & \textbf{55.02} & \textbf{64.79} \\
\bottomrule
\end{tabular}
\vspace{-0.4cm}
\end{table}

\noindent\textbf{Effect of MSG.}
Tab.~\ref{tab:msg_ablation} examines the contribution of MSG across different decoder architectures.
Incorporating MSG into the MLP decoder improves single-hop and overall Acc@1 by 1.27 and 0.46 percentage points, respectively.
For the CoCa decoder, MSG improves single-hop and multi-hop accuracy by 1.02 and 0.51 percentage points, respectively, although the reported overall accuracy slightly decreases.
These results suggest that the benefits of structured relational information vary across decoder architectures and question types.
\begin{table}[bhtp]
\vspace{-0.3cm}
\centering
\caption{\textbf{Effect of MSG.}
Ablation of MSG across different decoder architectures on City-3DQA under the city-wise split.
Acc@1 (\%) is reported.}
\label{tab:msg_ablation}
\small
\setlength{\tabcolsep}{5pt}
\renewcommand{\arraystretch}{1.12}
\begin{tabular}{@{}lcccc@{}}
\toprule
Decoder & MSG & Single-hop & Multi-hop & All \\
\midrule
MLP
& \xmark & 72.10 & 48.95 & 59.07 \\
MLP
& \cmark & 73.37 & 48.80 & 59.53 \\
\midrule
CoCa
& \xmark & 74.86 & 53.60 & \textbf{62.89} \\
CoCa
& \cmark & \textbf{75.88} & \textbf{54.11} & 62.62 \\
\bottomrule
\end{tabular}
\vspace{-0.4cm}
\end{table}

\noindent\textbf{Effect of visual feature fusion.}
Tab.~\ref{tab:fusion_ablation} compares feature concatenation and cross-attention using the same CoCa decoder and Qwen3.8-2B-Distill backbone.
Cross-attention improves overall Acc@1 from 63.81\% to 64.79\%, with gains of 1.04 and 0.93 percentage points on single-hop and multi-hop questions, respectively.
These consistent improvements suggest that cross-attention facilitates the integration of geometric information into language-based spatial reasoning.
\begin{table}[bhtp]
\vspace{-0.3cm}
\centering
\caption{\textbf{Effect of visual feature fusion.}
Comparison of concatenation and cross-attention with the same decoder backbone on City-3DQA under the city-wise split.
Acc@1 (\%) is reported.}
\label{tab:fusion_ablation}
\small
\setlength{\tabcolsep}{6pt}
\renewcommand{\arraystretch}{1.12}
\begin{tabular}{@{}lccc@{}}
\toprule
Fusion Method & Single-hop & Multi-hop & All \\
\midrule
Concatenation
& 76.34 & 54.09 & 63.81 \\
Cross-Attention
& \textbf{77.38} & \textbf{55.02} & \textbf{64.79} \\
\bottomrule
\end{tabular}
\vspace{-0.5cm}
\end{table}

\noindent\textbf{Effect of caption integration strategies.}
Tab.~\ref{tab:training_ablation} compares different strategies for incorporating scene captions into question answering.
Sequential and alternating training achieve overall Acc@1 ranging from 62.59\% to 63.20\%.
In contrast, integrating question-relevant captions achieves the highest overall Acc@1 of 64.79\%, outperforming the strongest alternative by 1.59 percentage points.
It also yields the best single-hop and multi-hop accuracies of 77.38\% and 55.02\%, respectively.
These results suggest that question-guided caption retrieval provides complementary semantic and relational context for urban 3D spatial reasoning.

\begin{table}[bhtp]
\vspace{-0.4cm}
\centering
\caption{\textbf{Effect of caption integration strategies.}
Comparison of sequential, alternating, and retrieval-based caption integration on City-3DQA under the city-wise split.
Acc@1 (\%) is reported.}
\label{tab:training_ablation}
\small
\setlength{\tabcolsep}{4pt}
\renewcommand{\arraystretch}{1.12}
\begin{tabular}{@{}lccc@{}}
\toprule
Strategy & Single-hop & Multi-hop & All \\
\midrule
QA $\rightarrow$ Caption
& 75.19 & 52.81 & 62.59 \\
Caption $\rightarrow$ QA
& 75.30 & 53.66 & 63.12 \\
Alternating Caption and QA
& 76.18 & 53.13 & 63.20 \\
Retrieved Caption + QA
& \textbf{77.38} & \textbf{55.02} & \textbf{64.79} \\
\bottomrule
\end{tabular}
\vspace{-0.4cm}
\end{table}

Further ablations on decoder architectures and pre-training configurations are provided in the \textbf{\textit{Supp}}.

\subsection{Qualitative Analysis}
Fig.~\ref{fig:qualitative} presents qualitative comparisons on City-3DQA under the city-wise split.
For multi-hop questions, 3D CoCa v3 correctly resolves functional relationships and relative spatial proximity while accurately counting buildings in single-hop questions.
Our method correctly answers all four illustrated questions, highlighting its effectiveness in geometric understanding and relational reasoning in unseen urban scenes.
Additional qualitative results are provided in the \textbf{\textit{Supp}}.

\section{Conclusion}
In this paper, we presented 3D CoCa v3, a contrastive and generative framework for urban 3D scene understanding.
Our framework employs an enhanced vision-language architecture with multi-scale geometric representations and language-model-based decoding.
We further introduced MSG-based contrastive learning to incorporate structured spatial relationships into 3D-language representations, together with City3D-9.6K, a multi-view urban dataset providing aligned visual-geometric observations and hierarchical language annotations.
Extensive experiments on City-3DQA demonstrated the effectiveness of our approach, achieving 88.35\% and 64.79\% overall top-1 accuracy on the sentence-wise and city-wise splits, outperforming 3D CoCa v2 by 20.88 and 3.57 percentage points, respectively.
Moreover, 3D CoCa v3 achieved the best performance on GeoEval3D.
These results highlight the effectiveness of jointly modeling geometric structures and relational semantics for urban 3D scene understanding and spatial reasoning.

\section*{LLM Usage Statement}
The authors used OpenAI's ChatGPT (GPT) to assist with language editing and refinement, related work and citation preparation, code and implementation review, experimental design and result analysis, figure preparation, and synthetic data annotation. No LLM or AI tools were used to formulate research ideas, hypotheses, claims, or conclusions. The authors have reviewed and verified the content and take full responsibility for the submitted work.

{
    \small
    \bibliographystyle{ieeenat_fullname}
    \bibliography{main}
}

\input{X_suppl}

\end{document}

%% file: X_suppl.tex
\clearpage
\setcounter{page}{1}
\maketitlesupplementary

\setcounter{section}{0}
\renewcommand{\thesection}{\Alph{section}}
\setcounter{table}{0}
\renewcommand{\thetable}{S\arabic{table}}
\setcounter{figure}{0}
\renewcommand{\thefigure}{S\arabic{figure}}

\section{Additional Ablation Study}
We provide additional ablation studies on City-3DQA under the city-wise split to examine the effects of decoder architectures, pre-training learning rates, and view sampling budgets.

\noindent\textbf{Effect of decoder architectures.}
Tab.~\ref{tab:decoder_arch_ablation} compares CapPa- and CoCa-based decoders with different language-model backbones.
Incorporating Qwen3-VL-2B-I improves overall Acc@1 from 60.95\% to 63.72\% for CapPa and from 62.89\% to 64.13\% for CoCa.
With Qwen3.8-2B-Distill and cross-attention, CoCa achieves the highest overall and multi-hop accuracies of 64.79\% and 55.02\%, respectively, outperforming its CapPa counterpart by 0.51 and 0.55 percentage points.
These results support our choice of decoder configuration.

\begin{table}[bhtp]
\vspace{-0.3cm}
\small
\centering
\caption{\textbf{Effect of decoder architectures.}
Comparison of CapPa and CoCa with different language-model backbones on City-3DQA under the city-wise split. Acc@1 (\%) is reported.
Qwen3.8 variants use cross-attention.}
\label{tab:decoder_arch_ablation}
\small
\setlength{\tabcolsep}{4pt}
\renewcommand{\arraystretch}{1.12}
\resizebox{0.95\linewidth}{!}{%
\begin{tabular}{@{}llccc@{}}
\toprule
Decoder & Backbone & Single-hop & Multi-hop & All \\
\midrule
CapPa & -- 
& 74.37 & 50.54 & 60.95 \\
CapPa & Qwen3-VL-2B-I
& \textbf{77.59} & 52.96 & 63.72 \\
CapPa & Qwen3.8-2B-Dis
& 76.92 & 54.47 & 64.28 \\
\midrule
CoCa & --
& 74.86 & 53.60 & 62.89 \\
CoCa & Qwen3-VL-2B-I
& 76.66 & 54.41 & 64.13 \\
CoCa & Qwen3.8-2B-Dis
& 77.38 & \textbf{55.02} & \textbf{64.79} \\
\bottomrule
\end{tabular}%
}
\vspace{-0.4cm}
\end{table}

\noindent\textbf{Sensitivity to pre-training learning rate.}
Tab.~\ref{tab:lr_ablation} evaluates the sensitivity of 3D CoCa v3 to the learning rate during contrastive pre-training.
Across learning rates ranging from $10^{-6}$ to $10^{-4}$, overall Acc@1 varies by only 0.23 percentage points, indicating limited sensitivity within the evaluated range.
We adopt $10^{-5}$ as the default learning rate, which achieves the highest overall and multi-hop accuracies of 64.79\% and 55.02\%, respectively.

\begin{table}[bhtp]
\vspace{-0.3cm}
\small
\centering
\caption{\textbf{Sensitivity to pre-training learning rate.}
Results with different learning rates for contrastive pre-training on City-3DQA under the city-wise split.
Acc@1 (\%) is reported.}
\label{tab:lr_ablation}
\small
\setlength{\tabcolsep}{6pt}
\renewcommand{\arraystretch}{1.12}
\begin{tabular}{@{}lccc@{}}
\toprule
Learning Rate & Single-hop & Multi-hop & All \\
\midrule
$10^{-6}$ & \textbf{77.58} & 54.47 & 64.56 \\
$10^{-5}$ & 77.38 & \textbf{55.02} & \textbf{64.79} \\
$10^{-4}$ & 77.32 & 54.98 & 64.74 \\
\bottomrule
\end{tabular}
\vspace{-0.4cm}
\end{table}

\noindent\textbf{Effect of the number of training views.}
Tab.~\ref{tab:num_views_ablation} evaluates different view sampling budgets during contrastive pre-training.
Sampling 32 views achieves the best performance across all three metrics, improving overall Acc@1 by 0.44 and 0.50 percentage points over the 8-view and 16-view settings, respectively.
The relatively small performance variations suggest that our framework can learn effective representations even with fewer sampled views.
We adopt 32 views per scene as the default configuration.

\begin{table}[bhtp]
\vspace{-0.3cm}
\centering
\caption{\textbf{Effect of the number of training views.}
Results with different numbers of sampled views per scene during contrastive pre-training on City-3DQA under the city-wise split.
Acc@1 (\%) is reported.}
\label{tab:num_views_ablation}
\small
\setlength{\tabcolsep}{6pt}
\renewcommand{\arraystretch}{1.12}
\begin{tabular}{@{}lccc@{}}
\toprule
\#Views & Single-hop & Multi-hop & All \\
\midrule
8  & 76.87 & 54.63 & 64.35 \\
16 & 76.90 & 54.51 & 64.29 \\
32 & \textbf{77.38} & \textbf{55.02} & \textbf{64.79} \\
\bottomrule
\end{tabular}
\vspace{-0.4cm}
\end{table}

\section{Additional Implementation Details}

We provide additional implementation details to facilitate reproducibility.
Both training stages are conducted on four NVIDIA A100 GPUs with 80 GB memory each.
The contrastive pre-training stage uses a batch size of 32 and runs for 6,000 epochs, while the multimodal decoder stage uses a batch size of 64.
During pre-training, 32 multi-view observations are sampled per scene, with RGB images and corresponding point maps resized to $224\times224$.
For downstream question answering, visual representations from 32 rendered views are extracted.
For the decoder, we use a 2B-parameter Qwen backbone with LoRA adaptation (rank 16 and $\alpha=32$) and an eight-head visual cross-attention module.
The pretrained language-model parameters remain frozen, while the LoRA parameters, visual projection layers, and cross-attention module are optimized.

\section{Training Dynamics}
We provide the training loss curves of both stages to illustrate the optimization behavior of 3D CoCa v3.

\begin{figure*}[htb]
    \centering
    \includegraphics[width=\linewidth]{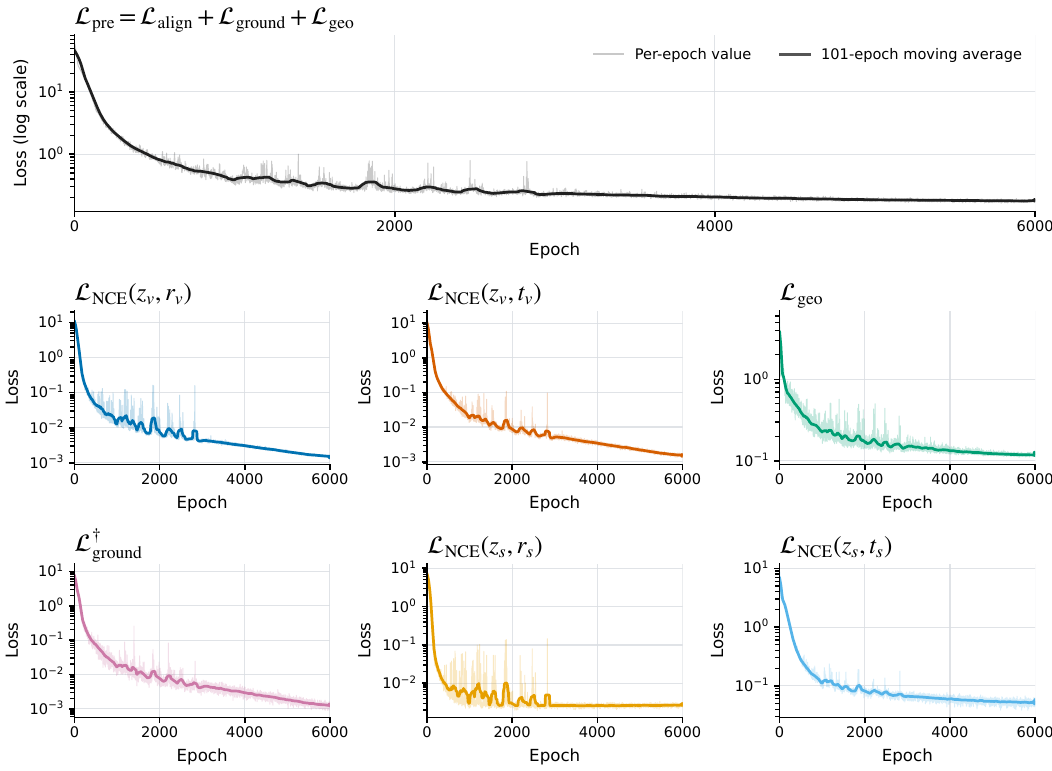}
    \caption{\textbf{Training curves of the contrastive pre-training stage.}
    The top panel shows the overall pre-training objective $\mathcal{L}_{\mathrm{pre}} = \mathcal{L}_{\mathrm{align}} + \mathcal{L}_{\mathrm{ground}} + \mathcal{L}_{\mathrm{geo}}$, while the remaining panels show the individual loss components.
    Light and dark curves represent per-epoch values and 101-epoch moving averages, respectively.
    All losses are plotted on a logarithmic scale.}
    \label{fig:pre_loss}
    \vspace{-0.4cm}
\end{figure*}

Fig.~\ref{fig:pre_loss} illustrates the optimization dynamics of the contrastive pre-training stage.
The overall objective and its individual components exhibit consistent decreasing trends throughout training.
Although transient fluctuations are observed during optimization, the moving-average curves gradually stabilize, indicating stable convergence of the geometric, grounding, and cross-modal alignment objectives.

\begin{figure}[htb]
    \centering
    \includegraphics[width=\linewidth]{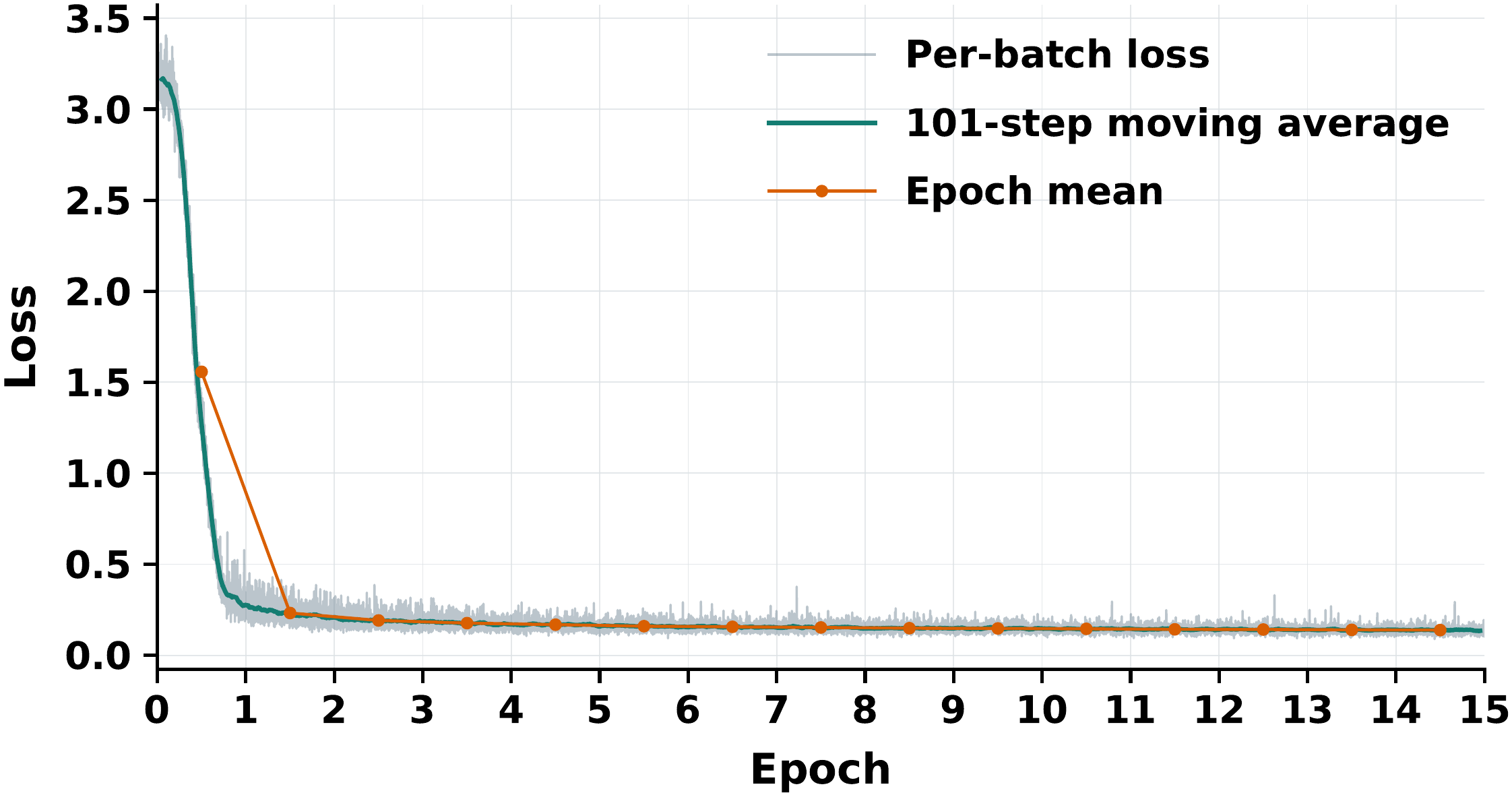}
    \caption{\textbf{Training loss of the decoder stage.}
    We report the per-batch training loss, its 101-step moving average, and the epoch-level mean over 15 training epochs.
    The loss decreases rapidly during the initial optimization stage and gradually stabilizes in subsequent epochs.}
    \label{fig:decoder_loss}
    \vspace{-0.4cm}
\end{figure}

Fig.~\ref{fig:decoder_loss} shows the training dynamics of the decoder stage over 15 epochs.
The training loss decreases rapidly during the initial updates, followed by a gradual reduction and stabilization in subsequent epochs.
The moving-average curve and epoch-level means exhibit consistent trends despite per-batch fluctuations, suggesting stable optimization of the decoder.

\begin{figure*}[htb]
    \centering
    \includegraphics[width=\linewidth]{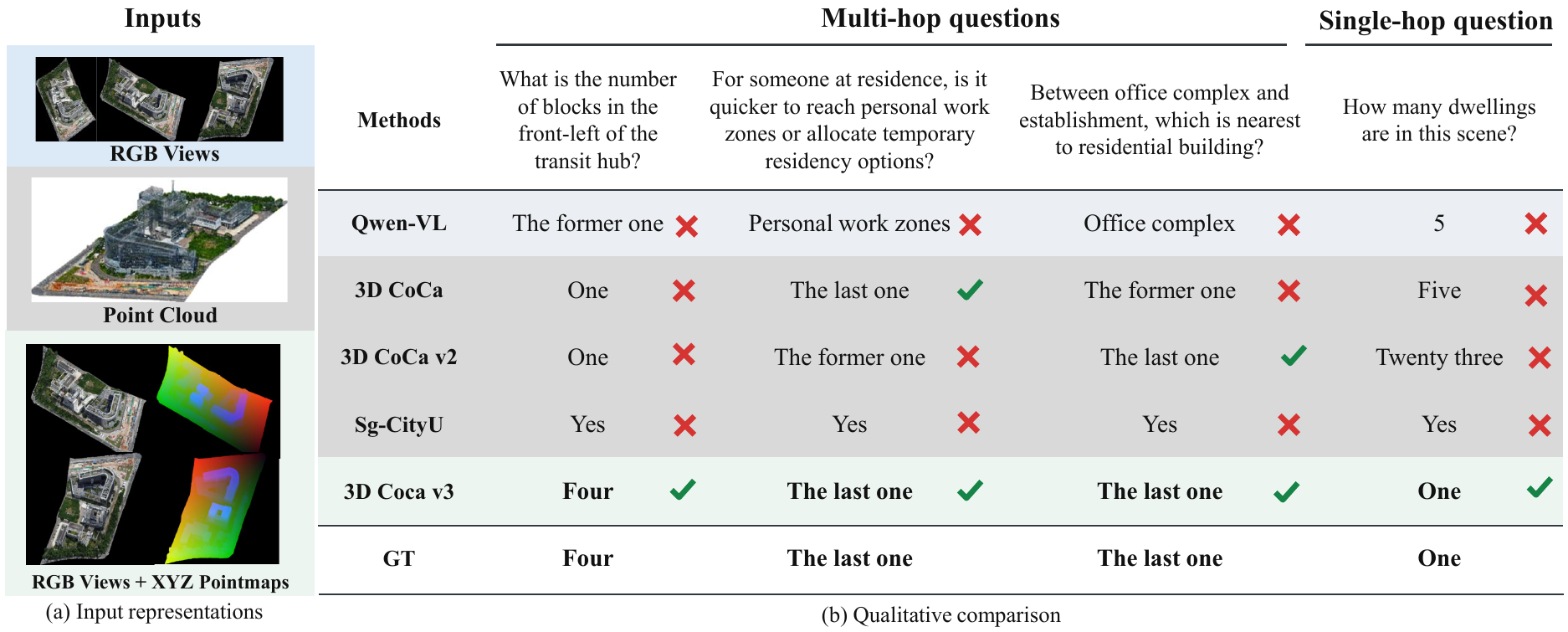}
    \caption{\textbf{Additional qualitative comparisons on City-3DQA under the city-wise split.}
    We compare 3D CoCa v3 with representative baselines on multi-hop and single-hop questions using multi-view RGB observations, point clouds, and corresponding XYZ point maps.
    Green checkmarks and red crosses indicate correct and incorrect predictions, respectively.
    Ground-truth (GT) answers are provided for reference.}
    \label{fig:qualitative2}
\end{figure*}

\begin{figure*}[htb]
    \centering
    \includegraphics[width=\linewidth]{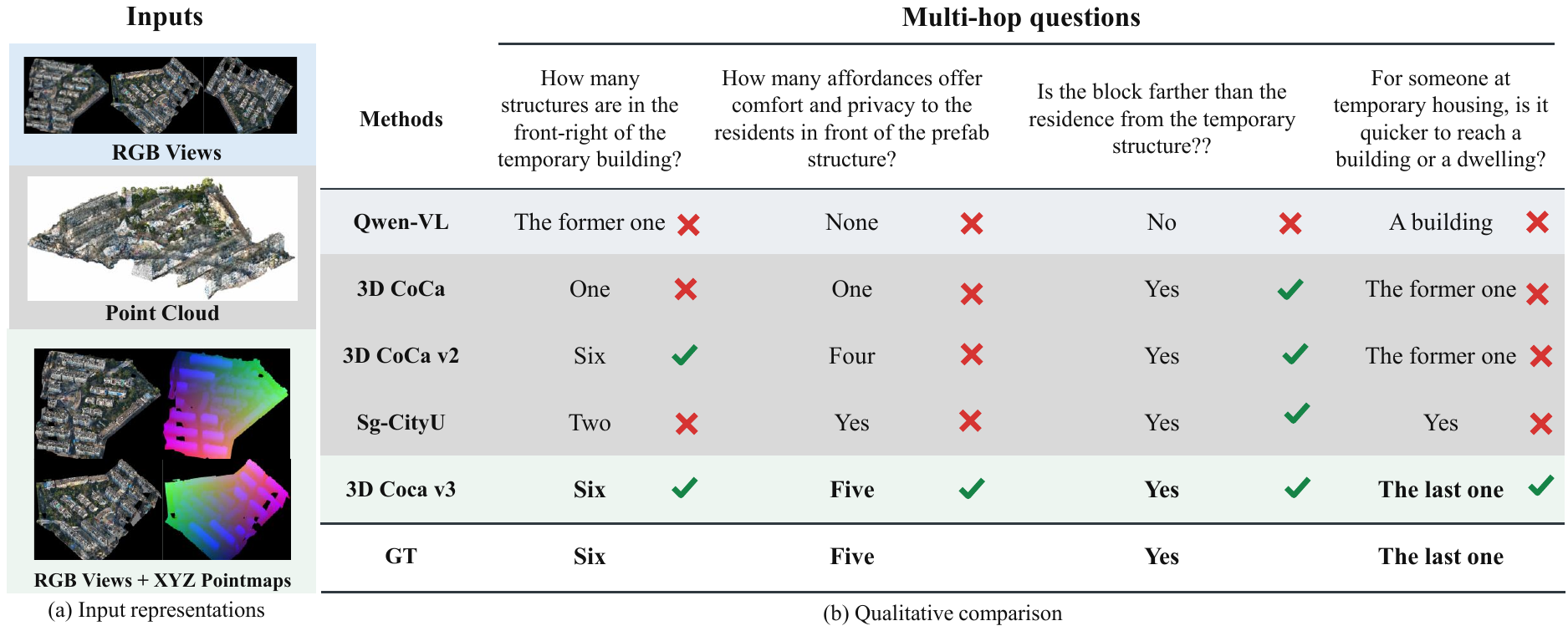}
    \caption{\textbf{Additional qualitative comparisons on City-3DQA under the sentence-wise split.}
    The examples focus on multi-hop questions involving object counting, functional relationships, and spatial reasoning.
    We compare predictions from different methods against the ground-truth (GT) answers.
    Green checkmarks and red crosses indicate correct and incorrect predictions, respectively.}
    \label{fig:qualitative3}
    \vspace{-0.6cm}
\end{figure*}

\section{Additional Qualitative Results}
Figs.~\ref{fig:qualitative2} and~\ref{fig:qualitative3} present additional qualitative comparisons on City-3DQA under the city-wise and sentence-wise splits, respectively.
The examples cover object counting, functional relationships, and relative spatial reasoning using multi-view geometric and visual observations.
3D CoCa v3 correctly answers all illustrated questions in both splits, while competing methods exhibit errors in several counting and relational reasoning tasks.
These results further illustrate the effectiveness of integrating multi-view geometric representations and structured relational information for urban 3D scene understanding.

\begin{figure*}[htb]
    \centering
    \includegraphics[width=\linewidth]{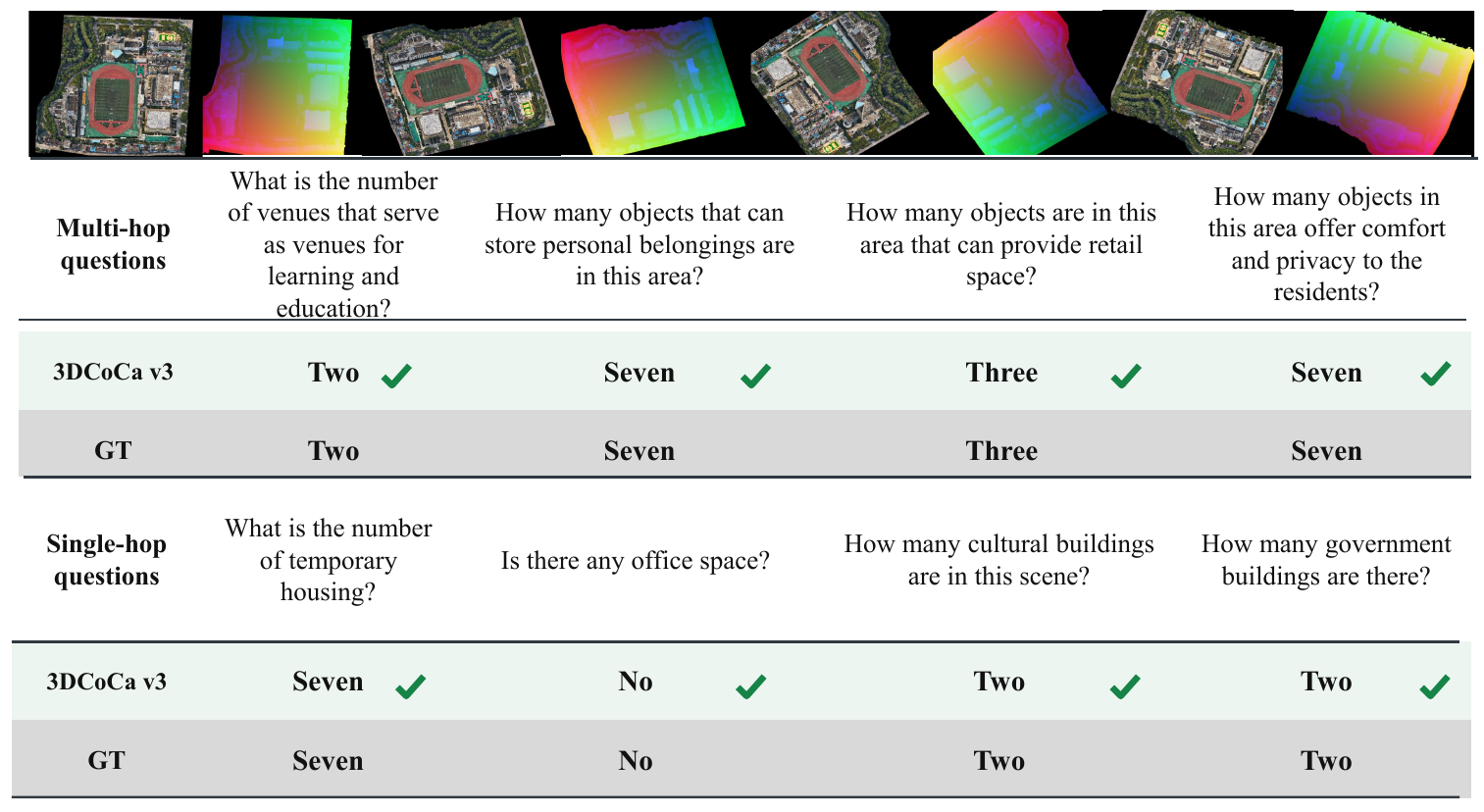}
    \caption{\textbf{Additional qualitative results on urban functional understanding.}
    Examples illustrate multi-hop and single-hop questions involving functional object counting, including educational, retail, residential, cultural, and governmental facilities.
    The predictions of 3D CoCa v3 consistently match the ground-truth (GT) answers.}
    \label{fig:qualitative4}
    \vspace{-0.4cm}
\end{figure*}

\begin{figure*}[htb]
    \centering
    \includegraphics[width=\linewidth]{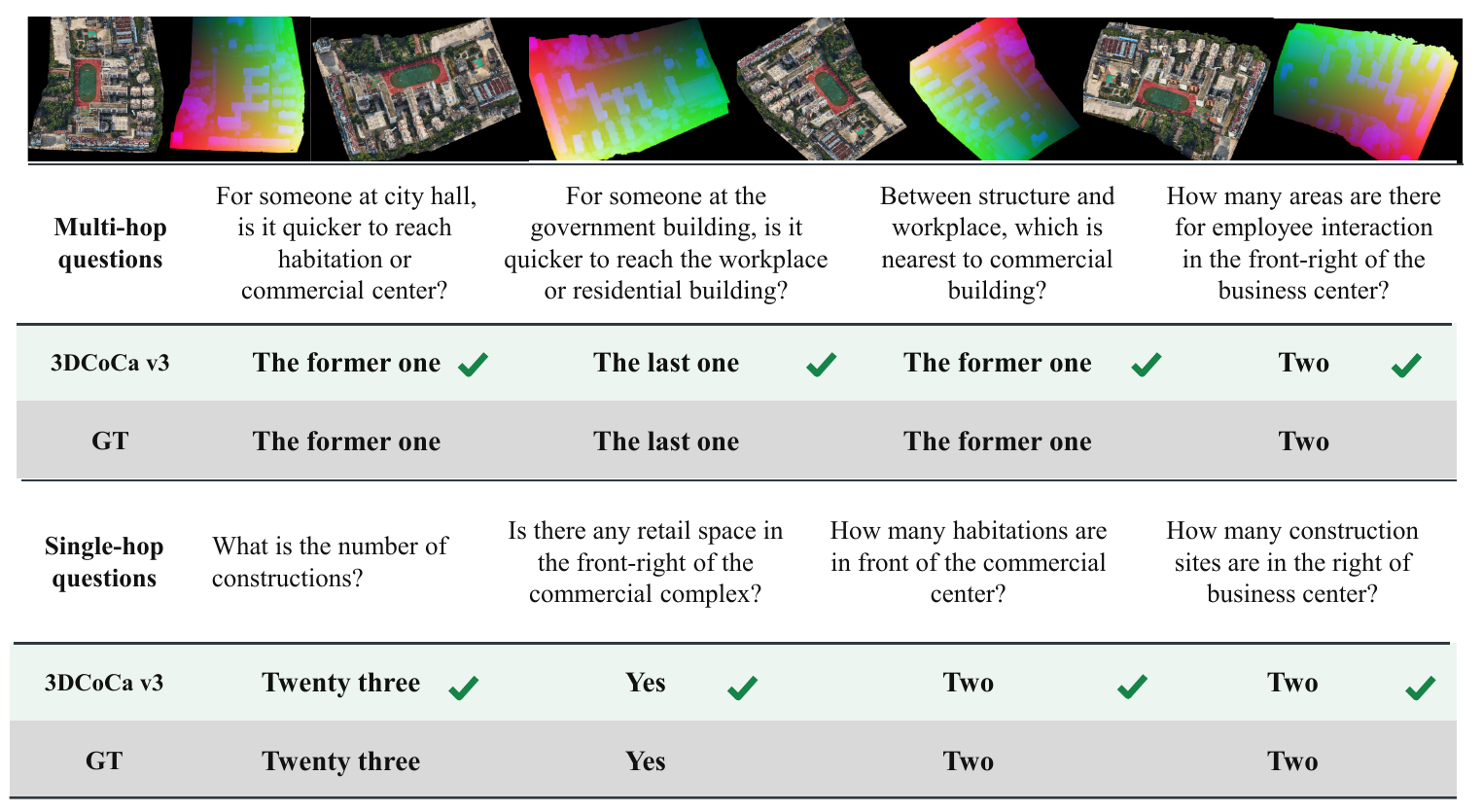}
    \caption{\textbf{Additional qualitative results on urban spatial relationships.}
    Examples demonstrate multi-hop reasoning about relative accessibility and spatial relationships among administrative, commercial, and residential areas, together with localized object counting and facility identification.
    3D CoCa v3 correctly answers all illustrated questions.}
    \label{fig:qualitative5}
    \vspace{-0.6cm}
\end{figure*}

\begin{figure*}[htb]
    \centering
    \includegraphics[width=\linewidth]{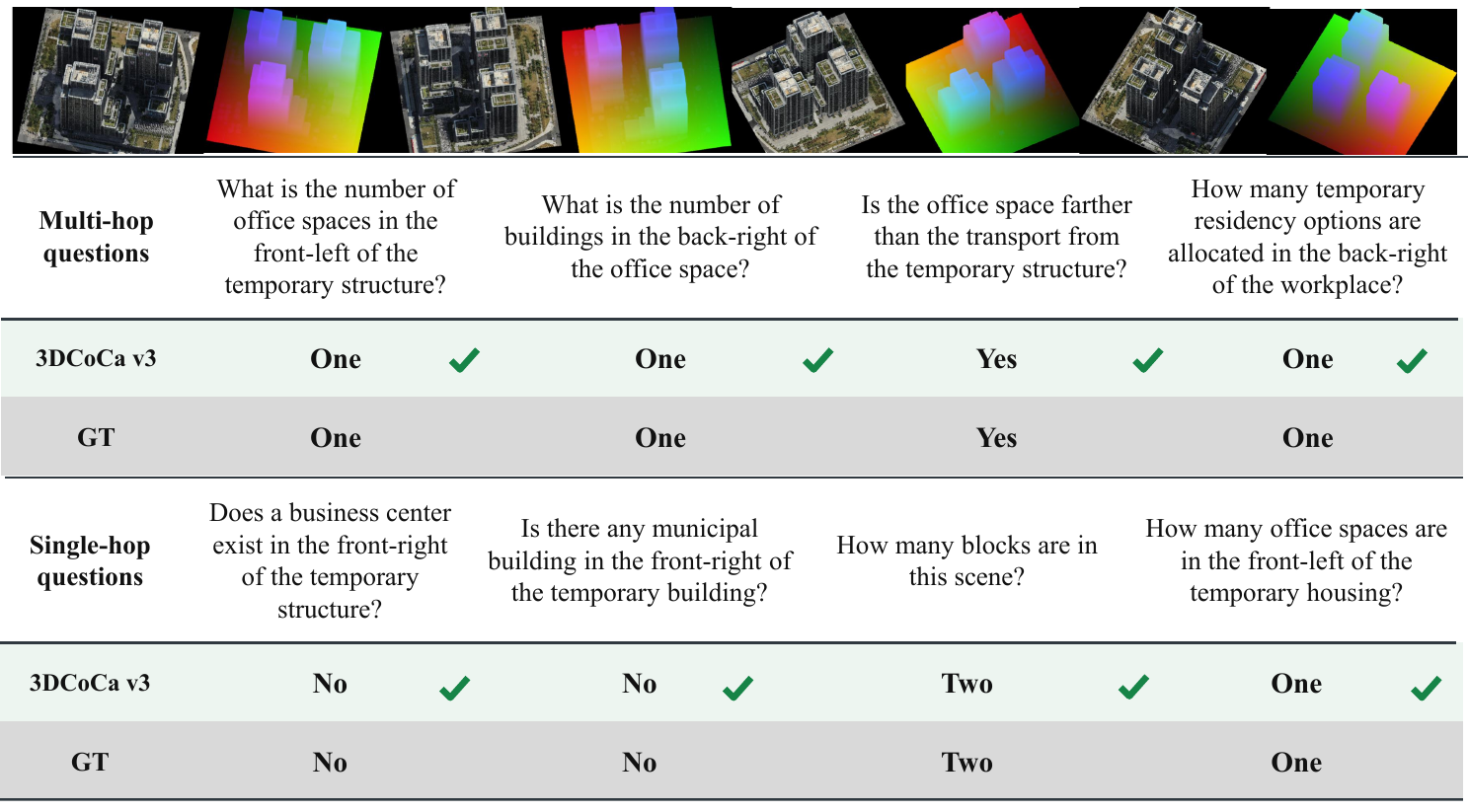}
    \caption{\textbf{Additional qualitative results on fine-grained spatial reasoning.}
    Examples involve spatially localized object counting, relative distance comparison, and facility existence in urban scenes containing office spaces and temporary structures.
    All predictions of 3D CoCa v3 are consistent with the ground-truth answers.}
    \label{fig:qualitative6}
    \vspace{-0.4cm}
\end{figure*}

\begin{figure*}[htb]
    \centering
    \includegraphics[width=\linewidth]{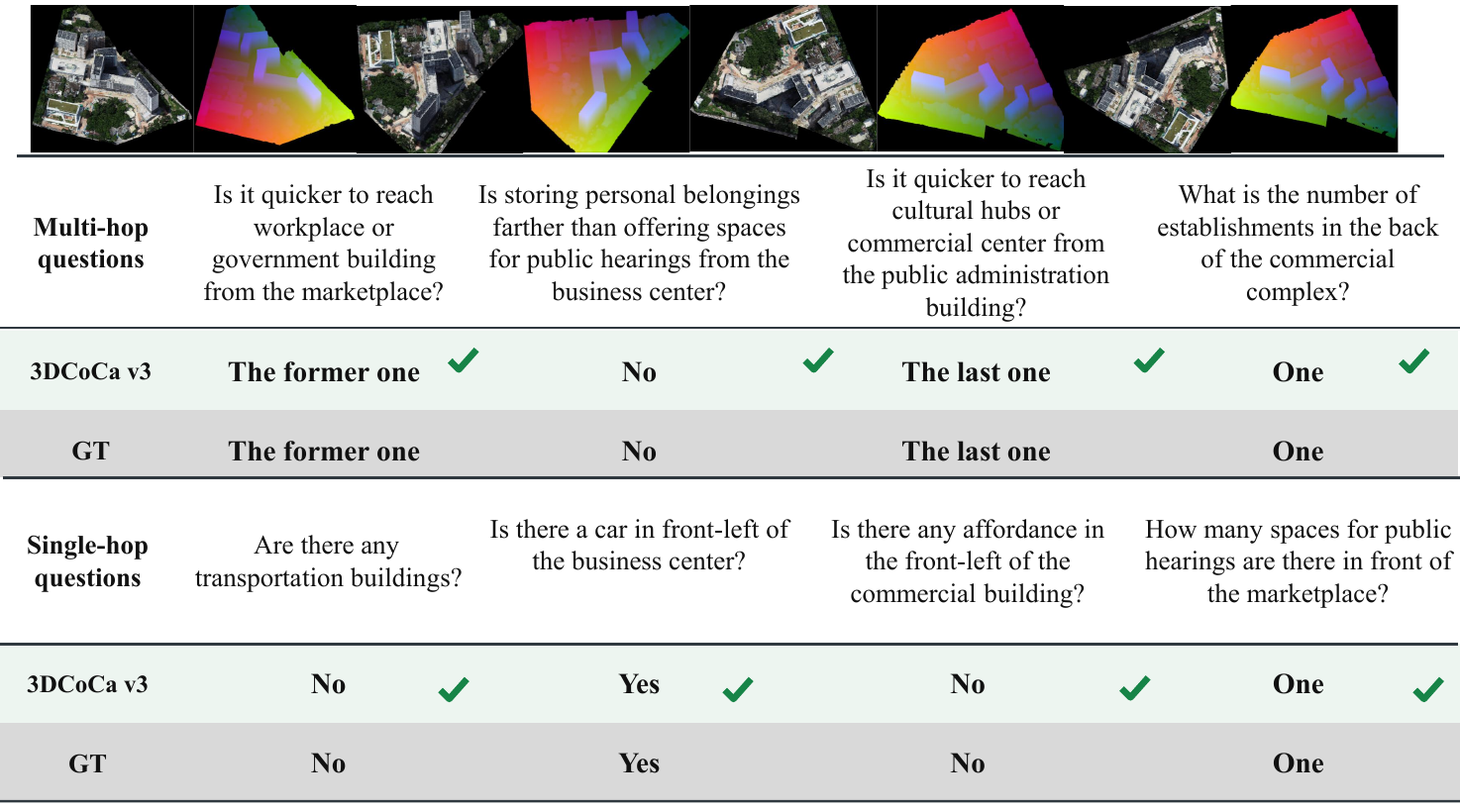}
    \caption{\textbf{Additional qualitative results on urban relational reasoning.}
    Examples illustrate relative accessibility, spatial proximity, object counting, and facility existence involving marketplaces, public administration buildings, and commercial areas.
    3D CoCa v3 correctly answers all illustrated single-hop and multi-hop questions.}
    \label{fig:qualitative7}
    \vspace{-0.6cm}
\end{figure*}

Figs.~\ref{fig:qualitative4}--\ref{fig:qualitative7} present additional predictions across diverse urban scenes.
The examples cover functional object counting, facility identification, relative spatial relationships, and accessibility reasoning, involving both single-hop and multi-hop questions.
3D CoCa v3 consistently produces predictions that match the ground-truth answers, including questions requiring fine-grained spatial localization and reasoning over multiple urban functional regions.
These results further demonstrate the effectiveness of our framework in integrating multi-view geometric observations and structured spatial information for urban scene understanding.

\section{Limiation and Future Work}
Despite the promising results, 3D CoCa v3 has several limitations.
First, our framework primarily focuses on static urban scene understanding and does not explicitly model temporal dynamics, such as moving objects and evolving spatial relationships.
Second, although City3D-9.6K provides complementary multi-view geometric and relational supervision, its scene coverage remains limited compared with the diversity of real-world urban environments.
Finally, while our method achieves strong performance in spatial reasoning and geometric understanding, fine-grained metric estimation and answer ranking still leave room for improvement.

Future work will explore extending 3D CoCa v3 to dynamic urban environments by incorporating temporal representations and spatiotemporal relational reasoning.

